\documentclass[10pt, a4paper]{article}

\usepackage{lrec-coling2024} 
\usepackage{multirow}
\usepackage{array}
\usepackage{graphicx}
\usepackage{caption}
\usepackage{subcaption}
\usepackage{color}
\usepackage{ragged2e}
\usepackage{enumitem}
\setlist[itemize]{leftmargin=0pt, align=parleft,left=0pt..1em}
\usepackage{float}
\usepackage{floatflt}

\title{\textbf{How Robust are the Tabular QA Models for Scientific Tables? A Study using Customized Dataset}}

\name{Akash Ghosh$^1$, B Venkata Sahith$^1$, Niloy Ganguly$^1$, Pawan Goyal$^1$, Mayank Singh$^2$}

\address{$^1$Indian Institute of Technology Kharagpur, India, $^2$Indian Institute of Technology Gandhinagar, India \\
         akashkgp@gmail.com, sminnu99@gmail.com, niloy@cse.iitkgp.ac.in,
        \\
         pawangiitk@gmail.com,
         singh.mayank@iitgn.ac.in \\}

\abstract{
Question-answering (QA) on hybrid scientific tabular and textual data deals with scientific information, and relies on complex numerical reasoning. In recent years, while tabular QA has seen rapid progress, understanding their robustness on scientific information is lacking due to absence of any benchmark dataset. To investigate the robustness of the existing state-of-the-art QA models on scientific hybrid tabular data, we propose a new dataset, ``SciTabQA'', consisting of 822 question-answer pairs from scientific tables and their descriptions. With the help of this dataset, we assess the state-of-the-art Tabular QA models based on their ability (i) to use heterogeneous information requiring both structured data (table) and unstructured data (text) and (ii) to perform complex scientific reasoning tasks. In essence, we check the capability of the models to interpret scientific tables and text. Our experiments show that ``SciTabQA'' is an innovative dataset to study question-answering over scientific heterogeneous data. We benchmark three state-of-the-art Tabular QA models, and find that the best F1 score is only 0.462. 
 \\ \newline  }

\begin{document}

\maketitleabstract

\section{Introduction}

\begin{figure*}[hbt!]
    \centering
    \includegraphics[width=15cm]{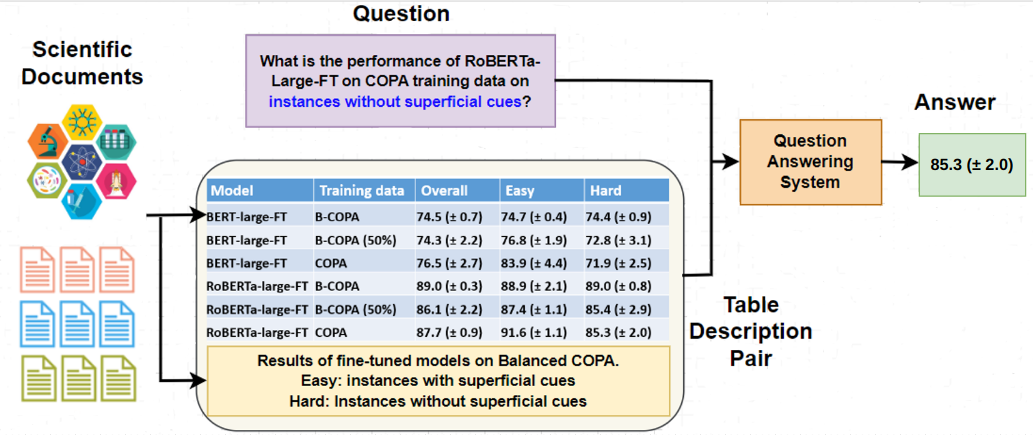}
    \caption{Scientific Hybrid Table Question Answering: for various questions, additional information from table captions, as well as table descriptions, may be required to come up with the appropriate answers. For instance, in the example, `instances without superficial cues' is understood only from the description.}
    \label{fig:p}
\end{figure*}

Question answering is a well-known task in NLP which focuses on answering a natural language question. It generally consists of an input passage from which the questions are to be answered. The input passage can be in the form of unstructured free-form text, for example Wikipedia passages as in the popular SQuAD question answering dataset ~\cite{rajpurkar-etal-2016-squad}, or structured data in the form of tables or databases ~\cite{krishnamurthy2017neural}. The existing datasets vary widely. Some may require arithmetic reasoning~\cite{lei-etal-2022-answering}, which is one of the important themes explored in our work.

Tabular QA involves answering natural language questions over tabular data. A number of datasets have been created for this task over the years, starting with WikiTableQuestions~\cite{pasupat-liang-2015-compositional} and WikiSQL~\cite{zhong2018seqsql}, which focus on Wikipedia tables. Extending tabular QA, hybrid QA focuses on answering questions with context as both table and text. ~\citet{jin2022survey} explored several domain-specific tabular and hybrid QA over a range of domains. Their exploration does not cover QA over scientific documents, which motivates the current work.

Scientific QA~\cite{lu2022learn,auer2023sciqa} involves answering questions based on scientific information. Hybrid scientific tabular QA involves answering questions from scientific tables and associated text. An illustrative hybrid scientific QA system is shown in Fig.~\ref{fig:p}, where the answer to the question depends on both the table and caption, showing the challenging nature of the task.

In this direction, we introduce our dataset ``SciTabQA''. The dataset consists of scientific table-description pairs and question-answer pairs over the \textit{Computer Science} (CS) domain. To the best of our knowledge, this is the first study on question-answering over tables and text in the scientific domain. Among the hybrid question-answering datasets,  HybridQA ~\cite{chen-etal-2020-hybridqa} and OTTQA~\cite{chen2021open} are based on Wikipedia data, and TATQA~\cite{zhu-etal-2021-tat}, FinQA~\cite{chen-etal-2021-finqa} and MultiHiertt~\cite{zhao-etal-2022-multihiertt} on financial data. In these datasets, the input tables are observed to be highly structured, as an example these tables avoid having nested headers. Financial datasets like FinQA and MultiHiertt in particular, have been annotated from FinTabNet ~\cite{zheng2021global} dataset, which contains annual reports from S\&P 500 companies, and thus are highly structured. 
When working with scientific papers and articles, the tabular data may not be structured, as evidenced from the Scigen~\cite{moosavi2021scigen} dataset, from which we obtain our dataset. This distinguishes our dataset, as there is a lack of hybrid datasets where the tables come in various formats. We created the dataset accommodating different table structures, as well as varying types of questions which require methods ranging from cell selection to numerical reasoning to answer. The dataset design, annotation and statistics are described in detail in Section 3.

In Section 6, we benchmark three state-of-the-art tabular pre-trained models, TAPAS~\cite{herzig-etal-2020-tapas}, TAPEX~\cite{liu2022tapex} and OmniTab~\cite{jiang-etal-2022-omnitab}. We observe that OmniTab performs the best on the ``SciTabQA'' dataset. Surprisingly, we find that adding caption and description information with the table degrades the overall performance. We then analyze the performance of the models on the individual tags and see that adding captions and descriptions helps only for the difficult questions, which need extra information. We also analyze the effect of truncation on the models.

The dataset and code are publicly available in Github. \footnote{\url{https://github.com/Akash-ghosh-123/SciTabQA}}

\section{Problem formulation}
Formally, each table instance in SciTabQA consists of a tuple of table T (which can further contain some rows and columns), caption c and description d. The tables, captions and descriptions are present in scientific articles, where caption consists of a few sentences present with the table and description is the text that refers and describes the table. We collectively represent the table instance as $(T, c, d)$, or shortened as $T^{'}$. Each data point consists of a table instance $T^{'}$, a question $q$, an answer $a$ and a tag \textbf{$\tau$}. 
The goal of the tabular question answering models is to generate the answer $a$ given the table $T^{'}$ and the question $q$, and can be formulated as 
\begin{equation}
   \arg \max_{a} p(a | q, T^{'})
\end{equation}

\section{Dataset}

Our dataset ``SciTabQA'' is a scientific question-answering (QA) dataset over hybrid textual and tabular data spanning 198 tables and 822 QA pairs. The data collection and human annotation processes are presented here.

\begin{figure*}[t]
    \centering
    \includegraphics[width=12cm]{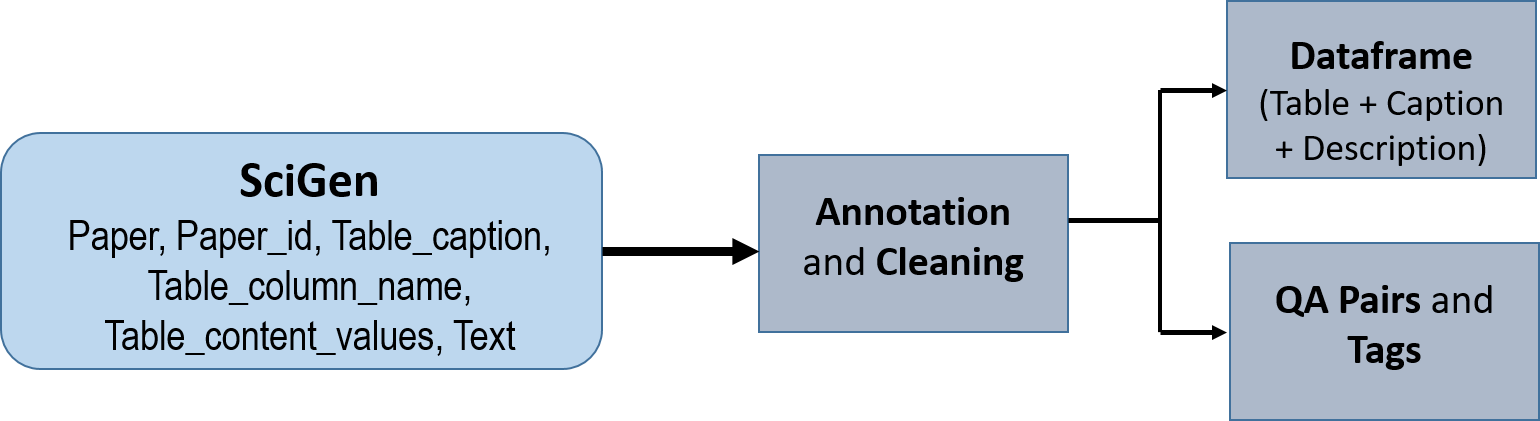}
    \caption{Data pre-processing and collection from SciGen to SciTabQA dataset.}
    \label{fig:pre_process}
\end{figure*}

\subsection{Data collection and preprocessing}

We collect the hybrid data from SciGen (\citet{moosavi2021scigen}) dataset, which consists of tables from scientific articles and their captions and descriptions, and was used to generate descriptions from the scientific tables.  
For our dataset creation, to obtain the correct tables, captions, and descriptions, we have used only those portions of training and
development sets of the SciGen dataset, which were fully annotated by expert annotators to have high quality annotations. We get a total of 220 table-description pairs, 200 from the train and 20 from development sets, extracted from ``Computation and Language'' articles.  Finally, our dataset consists of 822 question-answer pairs from 198 table-description pairs. 
The dataset creation pipeline is shown in Fig.~\ref{fig:pre_process}.

\subsection{Annotation Guidelines}

The annotation has been done by four undergraduate CS students with good domain knowledge in Computation and Language. 
Each annotator was responsible for annotating approximately 55 tables  
with question-answer pairs. The annotators were instructed to create between 4--5 question-answer pairs per table-description pair and were instructed to associate specific tags with each question-answer pair to be able to analyze the performance better. Specifically, a question-answer pair can have a single tag or multiple tags associated with it out of a total of 9 tags.  

The initial annotations from each of the annotators were re-evaluated and validated. In the original dataset, for 22 tables, the table contents were found to be difficult to interpret without going through the entire article, and these tables were dropped from the dataset.

To annotate question-answer pairs, guidelines were used to ensure that the questions have no ambiguity and are diverse. To ensure this, some examples of QA pairs for each tag were provided along with a set of general guidelines, namely, (i) Approximately a third of the questions should involve the selection of a cell or group of cells; (ii) One question per two tables should be answered with the help of corresponding text, which is the caption and description.

\vspace*{-2mm}
\subsection{Question tagging}
For our dataset, a total of 1,229 tags have been assigned for the 822 question-answer pairs. The tag statistics with their descriptions are shown in Table~\ref{table:sample-table1}. We illustrate the idea behind naming the tags, particularly Cell Selection (I) and Cell Selection (II). Cell Selection (I) questions can be answered from the content of a particular cell in the table, while Cell Selection (II) questions can be answered only by taking both table and text as context. We consider the question in Fig.~\ref{fig:p}, the tag for this question is Cell Selection (II), as answering it requires the knowledge that "instances without superficial cues" is equivalent to "hard", which comes from the caption.

\subsection{Inter-annotator agreement}
We have used four annotators with good domain knowledge. Each of the annotators annotated different parts of the dataset with no overlap, hence the concept of inter-annotator agreement is not directly applicable. Quality control of the annotations was made as they were checked by one person. In approximately 85\% of the cases, the original annotations were retained while appropriate changes were made in the remaining cases.

\begin{table*}[hbt!]
  
  \centering
  \begin{tabular}{|p{4cm}|p{5.5cm}|p{3.5cm}|p{2cm}|} \hline
  
    \textbf{Function} & \textbf{Description} & \textbf{Tags} & \textbf{Frequency}\\ \hline

    Cell selection operations & Simple cell selection from table & Cell Selection (I) & 386\\
    \hline
    
    Aggregate operations & Selecting rows and computing values by aggregating some rows/columns. & Sum/average/count & 211\\ \cline{3-4}
     & & Ordering/sorting & 137\\\cline{3-4}
    & & Selection by rank & 48\\ \hline

    Numerical operations &  Numerical operations based on arithmetic, scientific or logical knowledge. & Arithmetic operations (More complex like percentage etc.) & 194\\ \cline{3-4}
    & & Logical operations & 73\\ \cline{3-4}
    & & Scientific symbol operations & 55\\ \hline

    Others & Operations that involve reasoning from passages and might contain additional scientific context. & Cell Selection(II) (Both text and table as context) & 91\\ \cline{3-4} 
    & Includes questions that cannot be answered from table or text. & Negative answer & 34\\  
    \hline

  \end{tabular}
  \caption{Question tags grouped by broad types of questions, along with their frequency in the dataset.}
  \label{table:sample-table1}
\end{table*}

\begin{table*}[hbtp!]
    
    \label{results_table}
    \centering
    \begin{tabular}{|l|c|c|c|c|c|c|} \hline
     & \multicolumn{2}{c|}{\textbf{TAPAS}} &
    \multicolumn{2}{c|}{\textbf{TAPEX}} &
    \multicolumn{2}{c|}{\textbf{OmniTab}}
    \\\cline{2-7}
    &\textbf{EM} & \textbf{F1} & \textbf{EM} & \textbf{F1} &
    \textbf{EM} & \textbf{F1} \\ \hline

    Table & 0.352 & 0.429 & 0.357 & 0.406 & \textbf{0.397} & \textbf{0.462} \\ \hline
    Table + caption & 0.251 & 0.385 & 0.291 & 0.333 & \textbf{0.362} & \textbf{0.406} \\ \hline
    Table + caption + description & 0.118 & 0.154 & 0.231 & 0.272 & \textbf{0.296} & \textbf{0.365} \\ \hline
    
    \end{tabular}
    \caption{EM and F1 scores for various tabular models, while using only Table information, as well as adding caption and description. OmniTab performs the best overall, with just the table information. Adding extra information hurts all the models.}
    \label{table:results_table}
\end{table*}

\section{Baselines}

We use three pre-trained table question-answering models, TAPAS, TAPEX and OmniTab, as the baseline models to benchmark the performance of the SciTabQA dataset. These models have been used in standard TableQA tasks like WikiTableQuestions~\cite{pasupat-liang-2015-compositional} and WikiSQL~\cite{zhong2018seqsql}.

\begin{itemize}
    \item \textbf{TAPAS}~\cite{herzig-etal-2020-tapas}: TAPAS, or TAble PArSer is a weakly supervised table question answering model. TAPAS follows BERT~\cite{devlin-etal-2019-bert} encoder architecture
with additional row and column embeddings for encoding tabular structure. TAPAS is pre-trained from 6.2M table-text pairs from Wikipedia. TAPAS has a maximum token length of 512.

\item \textbf{TAPEX}~\cite{liu2022tapex}: TAPEX is pre-trained on tables by learning a executable SQL queries and their outputs. TAPEX addresses the data scarcity challenge via guiding
the language model to mimic a SQL executor on a diverse, large-scale and high-quality corpus. TAPEX has a maximum token length of 1024.

\item \textbf{OmniTab}~\cite{jiang-etal-2022-omnitab}: OmniTab is pre-trained on tables using both real and synthetic data. For pre-training, it uses retrieval to pair the tables with natural language. A SQL sampler randomly generates SQL queries from tables using a rule based method. Following this, synthetic questions answer pairs are generated from the SQL queries and their execution output. OmniTab has a maximum token length of 1024.

\end{itemize}

\section{Experiments}

For the pre-trained tabular question-answering models (baselines), we consider the following settings to explore if providing additional information such as caption and description can help:

\noindent\textbf{Table:} In this setting, we perform fine-tuning with only the table and question. 

\noindent\textbf{Table + caption:} In this setting, we perform fine-tuning with the table, question and caption. 

\noindent\textbf{Table + caption + description:} In this setting, we perform fine-tuning with the table, question, caption and description. 

The three baseline models work on only the table as input data. Hence, to incorporate the extra information from `table caption' as well as `table description', we append them to the question. So the whole context becomes question + caption + description + table. An important concern for large inputs is truncation of input data. In such cases, we avoid truncating the table, only truncating the caption and description instead.

\section{Results and Analysis}


\begin{table*}[!tbh]
  \centering
  \begin{tabular}{|l|c|c|c|c|c|c|} \hline
   \textbf{Tag} & \multicolumn{2}{c|}{\textbf{Table}} &
    \multicolumn{2}{c|}{\textbf{+ caption}} &
    \multicolumn{2}{c|}{\textbf{+ caption + description}}\\ \cline{2-7}
    & EM & F1 & EM & F1 & EM & F1 \\ \hline
    Cell Selection (I) & 0.632 & 0.647 & 0.529 & 0.537 & 0.382 & 0.4 \\
    Selection by rank & 0.367 & 0.482	&	0.368 & 0.461 & 0.315 & 0.392 \\
    Arithmetic operations & 0.154	& 0.154	& \textbf{0.168}	& \textbf{0.187}	& 0.111	& 0.122\\
    Cell Selection (II) & 0.228	& 0.26	&	\textbf{0.25} & \textbf{0.273}	& 	\textbf{0.243} & 0.26\\
    Logical operations & 0.147 & 0.185	&	0.138 & 0.172 &	0.129	& 0.168\\
    Ordering/sorting & 0.222 & 0.267	&	0.205 & 0.234	&	0.194 & 0.223\\
    Aggregate operations & 0.294 & 0.361	& 0.282	& 0.333	&	0.255	& 0.286\\
    Scientific symbol operations & 0.134	& 0.147	&	\textbf{0.142} & \textbf{0.155}	&	0.129 &  \textbf{0.153}\\
    Negative answer & 0.111	& 0.111	& 0.1 & 0.111	&	0.089 & 0.1\\ \hline
    Overall & 0.397 & 0.462 & 0.362 & 0.406 & 0.296 & 0.345\\ \hline
    
    \hline
 
  \end{tabular}
  \caption{Performance of OmniTab on various question tags, while using only Table, Table + caption, and Table + caption + description. Instances where additional information helps are highlighted in bold.}

  \label{table:sample-table-tags}
\end{table*}

\begin{table*}[!hbt]
    
    \centering
    \begin{tabular}{|l|c|c|c|c|c|c|} \hline
     & \multicolumn{2}{c|}{\textbf{TAPAS}} &
    \multicolumn{2}{c|}{\textbf{TAPEX}} &
    \multicolumn{2}{c|}{\textbf{OmniTab}}
    \\\cline{2-7}
    &\textbf{EM} & \textbf{F1} & \textbf{EM} & \textbf{F1} &
    \textbf{EM} & \textbf{F1} \\ \hline

    Table & 0.366 & 0.421 & 0.358 & 0.408 & \textbf{0.396} & \textbf{0.467} \\ \hline
    Table + caption & 0.317 & 0.368 & 0.310 & 0.372 & 0.362 & 0.406 \\ \hline
    Table + caption + description & 0.339 & 0.403 & 0.342 & 0.390 & 0.383 & 0.443 \\ \hline
    
    \end{tabular}
    \caption{EM and F1 scores for the pre-trained models with only the non-truncated examples used for caption and description.}
    \label{table:results_table_trunc}
\end{table*}

\begin{table*}[!hbt]
    
    \centering
    \begin{tabular}{|l|c|c|c|c|c|c|} \hline
  
     & \multicolumn{2}{c|}{\textbf{TAPAS}} &
    \multicolumn{2}{c|}{\textbf{TAPEX}} &
    \multicolumn{2}{c|}{\textbf{OmniTab}}
    \\\cline{2-7}
    &\textbf{EM} & \textbf{F1} & \textbf{EM} & \textbf{F1} &
    \textbf{EM} & \textbf{F1} \\ \hline

    Table & 0.188 & 0.223 & 0.204 & 0.251 & 0.232 & 0.279 \\ \hline
    Table + caption & 0.151 & 0.190 & 0.182 & 0.214 & 0.206 & 0.243 \\ \hline
    Table + caption + description & 0.091 & 0.137 & 0.130 & 0.179 & 0.165 & 0.203\\ \hline
    
    \end{tabular}
    \caption{EM and F1 scores for the models fine-tuned on the original WikiTableQuestions dataset.}
    \label{table:transfer_table}
\end{table*}

We have considered metrics used in standard question-answering, exact match (EM) and F1. The results for all the baseline models, under various settings, are shown in Table~\ref{table:results_table}. For TAPAS, TAPEX and OmniTab, we have fine-tuned on the training dataset. We observe that OmniTab gives the best results. However, the models generally do not provide high results on the proposed dataset, indicating the difficult nature of this dataset. Surprisingly, adding extra information in form of caption and description decreases the performance of all the models. We discuss it in detail below.

\subsection{Adding caption and description}

To understand the effect of adding caption and description, we further analyse the performance for different tags for OmniTab (the best performing model) in Table~\ref{table:sample-table-tags}, for the three settings. Interestingly, we observe that, for questions which require both textual and tabular information to answer, adding captions and descriptions have helped. For questions with tag Cell selection (II) adding caption and description has helped. The Exact Match (EM) for table-only is 0.228 and F1-score is 0.26. Adding caption increases the EM by 9.65\% and F1-score by 5\%. Adding caption and description increases the EM by 6.58\% and the F1-score remains the same. Hence, for questions where caption and description are important, we have found that performance increases as expected.

We observe that the questions that only require table information, e.g., Cell selection (I), aggregate operations and ordering/sorting, suffer the most in the (table + caption) and (table + caption + description) settings. We hypothesize that this may be due to the fact that adding caption and description actually adds more noise in the input.

\subsection{Truncation statistics}

Truncation of the input data is another major issue affecting the performance of the models. From table~\ref{table:truncated_table}, we observe that for TAPAS, around a third of inputs are truncated when caption and description are added. To understand the difference in performance, we run the experiments on only non-truncated examples in table~\ref{table:results_table_trunc}. The performance of TAPAS improves substantially in the (table + caption) and (table + caption + description) settings, and is almost similar to TAPEX. Thus, the relatively severe performance drop of TAPAS can be almost completely explained by the effect of truncation. 

\begin{table}[H] 

    \centering
    \begin{tabular}{|p{2.3cm}|c|c|c|} \hline
     & \textbf{TAPAS} & \textbf{TAPEX} & \textbf{OmniTab}
     \\ \hline
    Table & 8.04\% & 0\% & 0\%\\ \hline
    Table + caption & 9.55\% & 0\% & 0\%\\ \hline
    Table + caption + description & 33.67\% & 6.03\% & 6.03\% \\ \hline    
    \end{tabular}
    \caption{Proportion of inputs truncated for each of the models TAPAS, TAPEX and OmniTab.}
    \label{table:truncated_table}
\end{table}

\subsection{Transfer learning TableQA tasks}

The TAPAS, TAPEX and OmniTab models have been fine-tuned on WikiTableQuestions dataset, we checked the results for directly inferring the fine-tuned checkpoints on our test set. From table~\ref{table:transfer_table}, we observe that the results are much poorer, and thus training on our dataset improves the performance of the models.

\vspace*{-3mm}
\section{Conclusion}

We prepared the dataset SciTabQA and benchmark on pre-trained table QA models as well as hybrid QA models. It proves to be challenging for state-of-the-art table question-answering models as well as Hybrid question-answering models. This shows that scientific table question-answering, which is an important part of understanding scientific articles, needs better models.

\section{Limitations}
Some limitations of present work include the small size of the dataset, and the focus on a narrow domain within Computer Science. We plan to check if the findings also generalize to other domains. Also, for our dataset, we had the ground truth information available. It will be good to study if the model would still perform the same in the absence of ground truth caption and description information.

\nocite{*}
\section*{Bibliographical References}\label{sec:reference}

\bibliographystyle{lrec-coling2024-natbib}
\bibliography{lrec-coling2024-example}

\label{lr:ref}
\bibliographystylelanguageresource{lrec-coling2024-natbib}
\bibliographylanguageresource{languageresource}
\newpage

\end{document}